# Exploiting Domain Transferability for Collaborative Inter-level Domain Adaptive Object Detection

Mirae Do[a,∗], Seogkyu Jeon[b,∗], Pilhyeon Lee[b,∗], Kibeom hong[b], Yu-seung Ma[c,∗∗] and Hyeran Byun[a,b,∗∗]

[a]*Department of Artificial Intelligence, Yonsei University, Seoul, Korea*
[b]*Department of Computer Science, Yonsei University, Republic of Korea*
[c]*Electronics and Telecommunications Research Institute (ETRI), Daejeon, Korea*

ABSTRACT

Domain adaptation for object detection (DAOD) has recently drawn much attention owing to its capability of detecting target objects without any annotations. To tackle the problem, previous works focus on aligning features extracted from partial levels (e.g., image-level, instance-level, RPN-level) in a two-stage detector via adversarial training. However, individual levels in the object detection pipeline are closely related to each other and this inter-level relation is unconsidered yet. To this end, we introduce a novel framework for DAOD with three proposed components: Multi-scale-aware Uncertainty Attention (MUA), Transferable Region Proposal Network (TRPN), and Dynamic Instance Sampling (DIS). With these modules, we seek to reduce the negative transfer effect during training while maximizing transferability as well as discriminability in both domains. Finally, our framework implicitly learns domain invariant regions for object detection via exploiting the transferable information and enhances the complementarity between different detection levels by collaboratively utilizing their domain information. Through ablation studies and experiments, we show that the proposed modules contribute to the performance improvement in a synergic way, demonstrating the effectiveness of our method. Moreover, our model achieves a new state-of-the-art performance on various benchmarks.

## 1. Introduction

Object detection is one of the central problems in computer vision with a wide range of applications, aiming for precise localization and classification of objects in images. Thanks to the remarkable advent of deep neural networks, numerous object detection models (Liu, Anguelov, Erhan, Szegedy, Reed, Fu and Berg, 2016; Ren, He, Girshick and Sun, 2015) based on convolutional neural networks (CNNs) have successfully improved the performance using a large amount of labeled data (Lin, Maire, Belongie, Hays, Perona, Ramanan, Dollár and Zitnick, 2014). Nevertheless, they cannot guarantee their strong performance when facing data from other domains that are different from one for training due to the domain shift (Gopalan, Li and Chellappa, 2011). Suppose a detection model trained on sunny scene images for autonomous driving. When it encounters unseen domains, such as snowy and rainy scenes, its detection performance would be significantly degraded due to the large domain gap including camera setup, illumination, and object appearances. To handle this challenge, researchers have explored domain adaptive object detection (DAOD) mostly in the unsupervised setting (Chen, Zheng, Ding, Huang and Dou, 2020; Chen, Li, Sakaridis, Dai and Van Gool, 2018; Saito, Ushiku, Harada and Saenko, 2019; Shan, Lu and Chew, 2019; Xiong, Ye, Zhang, Gan and Hou, 2021), whose goal is to effectively transfer knowledge learned from a labeled source domain to an unlabeled target domain.

Since the seminal work of unsupervised domain-adaptive object detector (*i.e.*, DA-faster (Chen et al., 2018)), several approaches (Chen et al., 2020; He and Zhang, 2020; Kim, Choi, Kim and Kim, 2019; Nguyen, Tseng and Shuai, 2020; Rodriguez and Mikolajczyk, 2019; VS, Gupta, Oza, Sindagi and Patel, 2021; Zhu, Pang, Yang, Shi and Lin, 2019a) have tried to reduce the domain gap in the two-stage detection framework. Specifically, they handle the domain shift problem on three different levels in two-stage detector separately: (1) *image-level* features from the backbone network (Saito et al., 2019), (2) *RPN-level* regional features of pre-defined anchors from the region proposal network (RPN) (Zhao, Li, Xu and Lin, 2020), and (3) *instance-level* features after region of interests (ROI) operation (*e.g.*, pooling and aligning) for final object category classification and bounding box (Chen et al., 2018). It is well known that the modules in the two-stage object detection pipeline are closely related for detecting objects (Ren et al., 2015).

Nonetheless, existing studies (Chen et al., 2018; Nguyen et al., 2020; Saito et al., 2019; VS et al., 2021) overlook the fact that all the components in the object detection pipeline are interdependent and closely correlated to each other, which should be considered for effective adaptation. For instance, misaligned image-level features between the domains induce less discriminative foreground information, which in turn adversely affects the adaptation of the following detection levels (*i.e.*, RPN-level and instance-level).

∗Equal contributions
∗∗Co-corresponding authors
wwwdo109@yonsei.ac.kr (M. Do); jone9312@yonsei.ac.kr (S. Jeon); lph1114@yonsei.ac.kr (P. Lee); cha2068@yonsei.ac.kr (K. hong); ysma@etri.re.kr (Y. Ma); hrbyun@yonsei.ac.kr (H. Byun)
ORCID(s): 0000-0002-3082-3214 (H. Byun)

On the other hand, without RPN-level adaptation, the background regions of the target domain are likely to be misclassified as foreground, bringing about negative effects on instance-level adaptation. Motivated by these observations, we introduce a novel framework with three proposed modules carefully designed to collaboratively reduce the domain gap of individual levels in the two-stage detector pipeline.

For the *image-level* adaptation, we propose a *Multi-scale-aware Uncertainty Attention* (MUA). It is designed to highlight transferable regions while suppressing non-transferable regions (*e.g.*, background) to facilitate the adaptation of the subsequent levels. In contrast to the existing works (Chen et al., 2020; Nguyen et al., 2020), our MUA aggregates multi-scale image-level features from the backbone for transferability estimation. By incorporating both local and global information, the multi-scale strategy allows for the stable adaptation performance across various object sizes. This proposed module has two advantages: 1) transferring the reliable regions which focus on the foreground objects with the uncertainty attention map and 2) incorporating both local and global information with multi-scale strategy for various object sizes.

For the *RPN-level* adaptation, we propose a *Transferable Region Proposal Network* (TRPN) that effectively reduces the domain gap. Without the RPN-level adaptation, the source-biased knowledge leads to low objectness scores even for informative region proposals of the target domain. Our TRPN considers not only the objectness score but also the transferability of every anchor box, allowing the preservation of informative target regions for precise object detection. The underlying domain gap in RPN-level features from different anchors remains largely under-explored. The source-biased knowledge of RPN leads to low objectness scores even for informative region proposals of the target domain, which are therefore discarded during the proposal selection process. To overcome this problem, we propose a *Transferable Region Proposal Network* (TRPN) that effectively reduces the domain gap with selected adaptive-proper region proposals, considering not only the objectness score but also the transferability of every anchor box.

For the *instance-level* adaptation, we introduce *Dynamic Instance Sampling* (DIS) to prevent the negative effect caused by inaccurate target proposals in early training steps. Our motivation here is that the object candidates of the target domain are often unreliable when there exists misalignment in either image- or RPN-level features. In this case, involving all the proposals for model training is unhelpful or probably degenerative. Therefore, we propose to adjust the number of region proposals in accordance with the image-level hardness score of the target domain and the discrepancy of RPN-level foreground probability between the domains. This proposed method overcomes the limitation of fixed-number strategy that is undesirable in the adaptation scenario in that the regional proposals of the target domain are unreliable, especially in the early training steps, which eventually disturbs the adaptation process.

**Table 1**
GLOSSARY OF SYMBOLS.

| Symbols | Descriptions |
|---|---|
| $X_s$ | Set of source images |
| $Y_s$ | Set of source labels |
| $X_t$ | Set of target images |
| $F$ | Image-level feature extractor (backbone) |
| $F_{rpn}$ | Encoder of the region proposal network |
| $G_{cls}$ | Foreground classifier of the region proposal network |
| $G_{box}$ | Bounding box regressor of the region proposal network |
| $l$ | Layer index of the backbone $F$ |
| $H_l$ | Height of the output feature from layer $l$ |
| $W_l$ | Width of the output feature from the layer $l$ |
| $C_l$ | Channel size of the output feature from the layer $l$ |
| $D_{img}^l$ | Image-level domain discriminator corresponding to the layer $l$ |
| $D_{fus}$ | Multi-scale-aware domain discriminator |
| $D_{rpn}$ | RPN-level domain discriminator |
| $D_{ins}$ | Instance-level domain discriminator |
| $D_{dis}$ | Auxiliary domain discriminator for the multi-scale features |
| $\mathcal{L}_{det}$ | Overall loss function for the detector |
| $\mathcal{L}_{rpn}$ | Overall loss function for the region proposal network |
| $\mathcal{L}_{cls}$ | Classification loss for the classifier of detection head |
| $\mathcal{L}_{reg}$ | Regression loss for the bounding box regressor of detection head |
| $\mathcal{L}_{dis}$ | Loss function for the auxiliary multi-scale domain discriminator |
| $\mathcal{L}_{adv}^{img}$ | Image-level domain adversarial loss function |
| $\mathcal{L}_{adv}^{fus}$ | Multi-scale-aware Domain adversarial loss function |
| $\mathcal{L}_{adv}^{rpn}$ | RPN-level domain adversarial loss function |
| $\mathcal{L}_{adv}^{ins}$ | Instance-level domain adversarial loss function |

With these components, we estimate the domain transferability of each level and utilize it as a guidance for improving the robustness of the following level. Finally, our proposed frameworks enhance the transferability of the model against domain shifts while maintaining the discriminative ability on the object detection task. Through extensive experiments on various cross-domain detection benchmarks (*e.g.*, Cityscapes (Cordts, Omran, Ramos, Rehfeld, Enzweiler, Benenson, Franke, Roth and Schiele, 2016), KITTI (Geiger, Lenz, Stiller and Urtasun, 2013), SIM10K (Johnson-Roberson, Barto, Mehta, Sridhar, Rosaen and Vasudevan, 2016), BDD100K (Yu, Xian, Chen, Liu, Liao, Madhavan and Darrell, 2018)), we validate the superiority of our proposed method compared to the existing state-of-the-art works. Furthermore, a series of ablation studies clearly demonstrate the effectiveness of the proposed components.

## 2. Related work
### 2.1. Unsupervised Domain Adaptation

Unsupervised domain adaptation (UDA) can be deemed as a sub-problem of transfer learning, aiming to distill knowledge of models trained on a labeled source domain to an unlabeled target domain, where the two domains share the same task. To date, UDA has been investigated for various visual applications such as image classification (Chen, Wang, Long and Wang, 2019b; Ganin, Ustinova, Ajakan, Germain, Larochelle, Laviolette, Marchand and Lempitsky, 2016; Lee, Kim, Kim and Jeong, 2019; Yang, Xia, Ding and Ding, 2020), semantic segmentation (Chang, Wang, Peng and Chiu, 2019; Kim and Byun, 2020; Sankaranarayanan, Balaji, Jain, Lim and Chellappa, 2018; Tsai, Hung, Schulter, Sohn, Yang and Chandraker, 2018), and person re-identification (Fu, Wei, Wang, Zhou, Shi and Huang, 2019; Wei, Zhang, Gao and Tian, 2018; Zhai, Lu, Ye, Shan, Chen, Ji and Tian, 2020). Existing approaches focus on aligning the marginal feature distributions between source and target domains by matching the high-order statistics of feature distribution (Long, Cao, Wang and Jordan, 2015; Sun and Saenko, 2016; Tzeng, Hoffman, Zhang, Saenko and Darrell, 2014; Zellinger, Grubinger, Lughofer, Natschläger and Saminger-Platz, 2017) or using adversarial training (Chen, Xie, Huang, Rong, Ding, Huang, Xu and Huang, 2019a; Ganin et al., 2016; Pei, Cao, Long and Wang, 2018; Saito, Watanabe, Ushiku and Harada, 2018; Tzeng, Hoffman, Saenko and Darrell, 2017), generative adversarial networks (GANs) (Bousmalis, Silberman, Dohan, Erhan and Krishnan, 2017; Hoffman, Tzeng, Park, Zhu, Isola, Saenko, Efros and Darrell, 2018; Hu, Kan, Shan and Chen, 2018; Liu, Breuel and Kautz, 2017; Russo, Carlucci, Tommasi and Caputo, 2018), and self-training with pseudo-labels (Zou, Yu, Kumar and Wang, 2018). However, most of UDA methods are mainly focused on learning transferable features for domain alignment without integrating the class discrimination.

Recently, DWL (Xiao and Zhang, 2021) identifies that the imbalance between feature alignment and feature discrimination losses during domain alignment learning is problematic. The loss imbalance problem is also tackled in object detection area, e.g., (Wang and Zhang, 2021) which proposed to determine weighting factors of the sub-tasks by each other. To handle the imbalance, DWL (Xiao and Zhang, 2021) propose dynamically weighted learning to balance between the domain alignment and the class discrimination losses. Similarly, recent works (Chen et al., 2019b; Li, Zhai, Luo, Ge and Ren, 2020; Xiao and Zhang, 2021) attempt to measure the degree of aligning the entire feature distribution between domains. Unlike aforementioned studies that focus on the UDA for image classification, we consider the inter-relation between components of the object detector for effective domain adaptation. Our proposed methods reduce the domain gap of object detector regarding the property of each level to achieve higher performance on the target domain.

## 2.2. Domain Adaptive Object Detection

Recently, UDA for the object detection task has drawn a lot of attention due to its various applications (Chen et al., 2020,1; Guan, Huang, Xiao, Lu and Cao, 2021; He and Zhang, 2019; Nguyen et al., 2020; Saito et al., 2019; Shan et al., 2019; Xie, Yu, Wang, Wang and Zhang, 2019; Xiong et al., 2021; Zhao et al., 2020). As a pioneering work, Domain Adaptive Faster R-CNN (DA-Faster) (Chen et al., 2018) reduces the domain gap in the two-stage detector framework (Ren et al., 2015). After that, several works (He and Zhang, 2019; Saito et al., 2019; Xie et al., 2019) reduce the domain gap using the hierarchical domain feature discriminator. SW-Faster (Saito et al., 2019) and (Xie et al., 2019) points out that conducting distribution alignment with high-level features can be problematic due to large domain gap such as distinct scene layouts. To tackle this, they propose to split the domain alignment into local alignment and global alignment using low-level and high-level features, respectively. Similarly, (He and Zhang, 2019) propose a hierarchical domain feature alignment module to boost training efficiency of DAOD. Recent main streams in DAOD studies are exploiting spatial attention maps (Chen et al., 2020; Nguyen et al., 2020; Zhao et al., 2020) and prediction uncertainty (Guan et al., 2021; Munir, Khan, Sarfraz and Ali, 2021; Nguyen et al., 2020) for effective domain adaptation.

In detail, HTCN (Chen et al., 2020) estimates the transferable regions both locally and globally, and utilize them to prevent the deterioration of discriminability in the target domain. MEAA (Nguyen et al., 2020) estimate transferable regions in the input image by measuring pixel-wise domain uncertainty and highlight the local features. (Guan et al., 2021) adopts the curriculum learning strategy for instance-level adaptation in accordance with the prediction entropy of region proposals. SSAL (Munir et al., 2021) estimates the prediction uncertainty from the feature extractor by applying Monte-Carlo dropout to capture the variations in both localization and confidence predictions. On the other hand, we measure the spatial prediction uncertainty map from the Image-level domain discriminator to give more attention on transferable regions.

Meanwhile, CT-DA (Zhao et al., 2020) signifies the importance of RPN-level alignment by pointing out that only image-level alignment cannot guarantee high-quality region proposals. Accordingly, they conduct the collaborative self-training between the region proposal network (RPN) and the region proposal classifier (RPC).

RDA (Huang, Guan, Xiao and Lu, 2021) propose a novel Fourier adversarial attacking (FAA) method to handle the overfitting problem in various unsupervised domain adaptation tasks (UDA). FAA perturbs the frequency component of the input image to mitigate overfitting of domain adaptation methods, resulting in enhanced robustness.

In addition, ATF (He, Zhang, Yang and Gao, 2021) handles the model collapse problem in DAOD which is caused from the parameter-shared backbone that harms the discriminability learned from the source-domain while aligning with the inaccurate target distribution. To this end, they alleviate the model collapse by generating target features from the ancillary net trained on the label-rich source data and transferring the domain-invariant information with the channel attention mechanism.

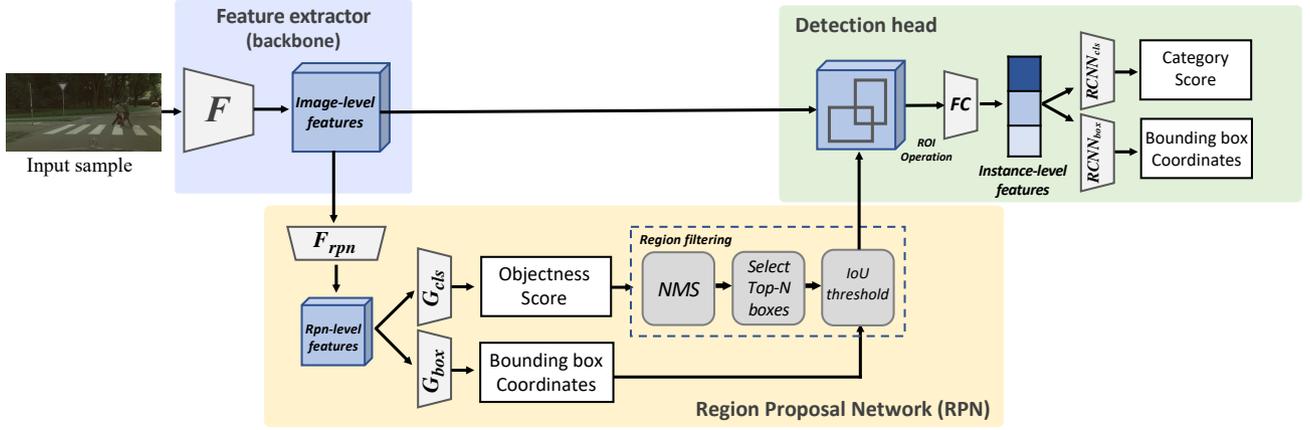

Figure 1: Overview of the two-stage detector framework.

Although the previous works achieve improvements in reducing the domain gap, they overlook the fact that all components in the object detection pipeline are indispensable and closely interrelated to each other. On the contrary, we propose novel components that collaboratively perform the adaptation in individual levels in order to fully exploit the complementary nature of the levels in the two-stage detector pipeline.

## 3. Proposed Method

In this section, we describe the proposed method. Our model follows the common two-stage detector pipeline but with three novel modules: Multi-scale-aware Uncertainty Attention (MUA), Transferable Region Proposal Network (TRPN), and Dynamic Instance Sampling (DIS). We first provide preliminaries and then describe each proposed module in detail. The symbols used in this paper are listed in Table 1.

### 3.1. Preliminaries

**Detection networks.** For the clarity of explanation, we briefly describe the two-stage detector framework, specifically Faster R-CNN (Ren et al., 2015), which is widely adopted as the baseline of domain adaptive object detection. As shown in Figure 1, Faster R-CNN is composed of an image-level feature extractor (*i.e.*, backbone), region proposal network (RPN), and the detection head. In detail, taking an image as input, the backbone $F$ extracts the *image-level features*. They are in turn fed into a $3 \times 3$ convolutional layer, *i.e.*, $F_{rpn}$, to obtain the *rpn-level features*. Given the rpn-level features, RPN predicts the objectness score and the bounding box coordinates for a set of region proposals by exploiting the pre-defined anchors with various sizes and aspect ratios. Afterwards, non-maximum suppression (NMS) is applied to remove highly overlapping region proposals and only the top-$N$ proposals are kept with regard to the objectness scores. To train the detection head, the proposals whose intersection-over-unions (IoUs) with ground-truths are higher than the threshold $\theta_H$ are labeled as foregrounds, while those with IoUs lying between $\theta_L$ and $\theta_H$ are considered as backgrounds.

We train the object detector on the fully-labeled source domain by minimizing the following loss function.

$$\mathcal{L}_{\text{det}} = \mathcal{L}_{\text{rpn}} + \mathcal{L}_{\text{reg}} + \mathcal{L}_{\text{cls}}, \qquad (1)$$

where $\mathcal{L}_{\text{rpn}}$, $\mathcal{L}_{\text{reg}}$, and $\mathcal{L}_{\text{cls}}$ denote the region proposal loss, the region regression loss, and the classification loss, respectively.

We train the detector on the fully-labeled source dataset, $\mathcal{D}_s = \{X_s, Y_s\}$ to obtain the discriminative knowledge for the source domain, where $X_s$ denotes a set of images $\{x_s\}$ and $Y_s$ is a set of the corresponding ground-truth labels $\{y_s\}$. Each label $y_s$ consists of bounding box coordinates and the object category of $x_s$. Our goal is to enhance the model performance on the target domain which is unlabeled ($\mathcal{D}_t = \{X_t\}$).

### 3.2. Multi-scale-aware Uncertainty Attention

The principle of domain adaptive object detection is extracting domain-invariant semantic information. Intuitively, descriptions of the same object regions in different domains are more semantically domain-invariant than those of background regions. Motivated by this, we aim to model spatial attention maps (Wang, Jiang, Qian, Yang, Li, Zhang, Wang and Tang, 2017) that can give guidance to indicate transferable regions, which are indistinguishable object regions in terms of the domain. To this end, we use the information theory (Chen et al., 2020; Nguyen et al., 2020) w.r.t. the domain discriminator to estimate the spatial attention map. In detail, the spatial attention map is derived by using the output entropy from the pixel-wise domain discriminator as follows.

$$\mathcal{H}(p_{i,l}) = -p_{i,l} \log p_{i,l}, \qquad (2)$$

where $p_{i,l} = D_p(f_{i,l})$ is the probability of the pixel-wise domain discriminator and $f_{i,l} \in \mathbb{R}^{C_l \times H_l \times W_l}$ is the feature of $i$-th image after the $l$-th layer. We estimate the spatial domain uncertainty $\mathcal{H}(p_{i,l}) \in \mathbb{R}^{H_l \times W_l}$ from the output domain probability $p_{i,l}$. Theoretically, $\mathcal{H}(p_{i,l})$ can be obtained regardless of the layer index $l$. If we derive $\mathcal{H}(p_{i,l})$ from the low-level features, it can provide domain-invariant structural details contributing to detecting small and distant objects (Nguyen et al., 2020; Yang, Choi and Lin, 2016). However, $\mathcal{H}(p_{i,l})$ may also regard the background regions (*e.g.*, road and the sky) to be transferable, and it will cause the degradation of detection performance. Meanwhile, the spatial uncertainty attention map derived from high-level features can reflect the contextual information, so backgrounds are better distinguished. Yet, it probably ignores small objects when evaluating the transferability due to the large receptive field (Yang et al., 2016). Therefore, we fuse the multi-scale features then estimate the spatial uncertainty attention map to minimize the shortcomings while harnessing all advantages.

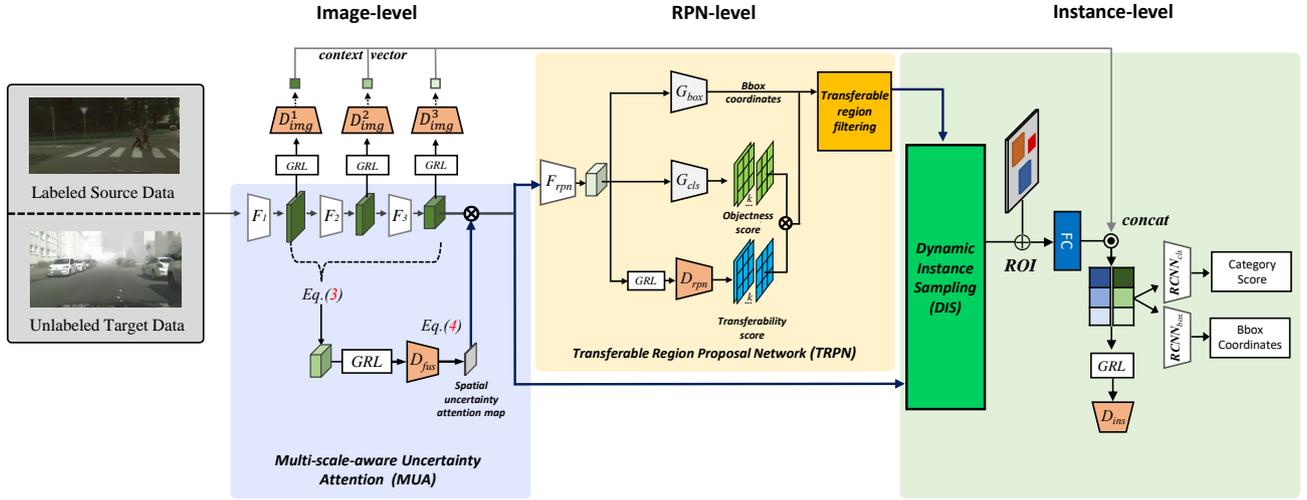

**Figure 2:** Overview of the proposed framework consisting of three components that perform domain alignment on individual levels: (1) image-level: Multi-scale-aware Uncertainty Attention (MUA), (2) RPN-level: Transferable Region Proposal Network (TRPN), and (3) instance-level: Dynamic Instance Sampling (DIS). $k$ denotes the number of anchors in RPN.

Specifically, we first re-scale the features from different layers to make them have the same spatial resolution with the feature of the last layer. Then, we concatenate them and embed the integrated feature into a latent space. The multi-scale feature fusion is formulated as:

$$M_i = \mathcal{F}\left[\mathcal{T}_l\left(f_{i,l}\right)\right]_{l=1}^{L}, \quad (3)$$

where $\mathcal{T}_l : \mathbb{R}^{C_l \times H_l \times W_l} \to \mathbb{R}^{C_L \times H_L \times W_L}$ denotes the bilinear interpolation function, $[\cdot]$ refers to the channel-wise concatenation operator, and $\mathcal{F}$ is a $1 \times 1$ convolutional layer for embedding. $L$ denotes index of the last layer in the feature extractor. With the multi-scale representation $M_i$, we calculate the Multi-scale-aware Uncertainty Attention (MUA) as follows.

$$E_i = \mathcal{H}\left(D_{fus}\left(M_i\right)\right), \quad (4)$$

where $D_{fus}$ is the multi-scale-aware domain discriminator of the fused feature $M_i$, while $E_i$ is the estimated spatial uncertainty attention map (MUA). The pixel-wise domain adversarial loss (Mao, Li, Xie, Lau, Wang and Paul Smolley, 2017; Saito et al., 2019) to train the multi-scale-aware domain discriminator is defined as follows.

$$\mathcal{L}_{adv}^{fus} = \frac{1}{n_s H_L W_L} \sum_{i=1}^{n_s} \sum_{w=1}^{W_L} \sum_{h=1}^{H_L} \left(D_{fus}\left(M_i^s\right)_{wh}\right)^2$$
$$+ \frac{1}{n_t H_L W_L} \sum_{i=1}^{n_t} \sum_{w=1}^{W_L} \sum_{h=1}^{H_L} \left(1 - D_{fus}\left(M_i^t\right)_{wh}\right)^2, \quad (5)$$

where $D_{fus}\left(M_i\right)_{wh}$ is the output of the the multi-scale-aware domain discriminator in each location and $n_s$ denotes the number of source examples. Note that we apply the gradient reversal layer (GRL) between the multi-scale-aware domain discriminator and the embedding layer $\mathcal{F}$ for the domain-invariant feature learning.

As the final step, we multiply the global feature with the spatial uncertainty attention map $\tilde{f}_{i,L} \leftarrow f_{i,L} \times E_i$. Consequently, the spatial uncertainty attention map unveils transferable regions from the local and global features, greatly enhancing the ability of representation on objects of various sizes. We further validate the efficacy of MUA through experimental analyses in Section 4.4.

In addition, to further enhance the domain invariance in image-level features, we design the layer-wise domain discriminators $\{D_{img}^l\}_{l=1}^{L}$. They are optimized to minimize the binary cross-entropy loss as follows.

$$\mathcal{L}_{adv}^{img} = -\sum_{i,l} \left(d_i \log p_{i,l} + (1 - d_i) \log(1 - p_{i,l})\right), \quad (6)$$

where $p_{i,l} = D_{img}^l(f_{i,l})$ is the probability of the domain discriminator and $\tilde{f}_{i,l}$ is the feature of $i$-th image from the $l$-th layer of backbone. Here $d_i$ denotes the domain label of the $i$-th image, with $d_i = 0$ if the image comes from the source domain, otherwise 1. The sign of the gradient is reversed when passing through the GRL layer to optimize the backbone network.

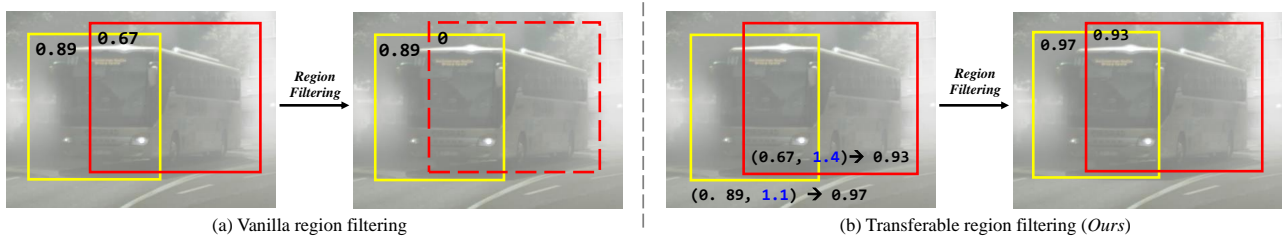

**Figure 3:** Comparison of the vanilla region filtering and our transferable region filtering. Two different box predictions for the *bus* are marked as red and yellow colors respectively, while the corresponding objectness and uncertainty scores are denoted in black and blue colors respectively. (a) Since the vanilla one filters detection boxes in terms of the objectness score only, an informative region (red) is discarded. (b) In contrast, our filtering method retains both regions by simultaneously considering domain entropy and objectness.

### 3.3. Transferable Region Proposal Network

In this section, we elaborate on the novel Transferable Region Proposal Network (TRPN), which alleviates the domain gap in RPN-level and generates foreground region proposals by considering both the objectness and the transferability.

#### 3.3.1. RPN-level domain adaptation

Previous studies (Chen et al., 2020,1; Nguyen et al., 2020; Saito et al., 2019; Xie et al., 2019) expect that aligning distributions of domains in image-level and instance-level is enough for RPN to operate properly on the target domain. However, the domain discrepancy still exists in the features from anchors of RPN module since learning the objectness prediction and the bounding box coordinates prediction is biased to the source domain. To address this problem, we propose an auxiliary domain discriminator $D_{rpn}$ which predicts the domain of region proposals produced from various anchors. The anchor-wise domain adversarial loss is defined as follows:

$$\mathcal{L}_{adv}^{rpn} = -\frac{1}{R} \sum_{i,r} \left( d_i \log p_{i,r}^{rpn} + (1 - d_i) \log \left(1 - p_{i,r}^{rpn}\right) \right). \tag{7}$$

Here, we denote the probability of the domain discriminator as $p_{i,r}^{rpn} = D_{rpn}(F_{rpn}(\tilde{f}_{i,L}))$, and $R \in \mathbb{R}^{k \times H \times W}$ is the number of region proposals using $k$ anchors. $r$ is the index of each region proposal in the $i$-th image. The domain labels $d_i$ is 0 if the image is from the source domain, otherwise 1. As other domain discriminators, GRL is applied between the RPN-level feature-extracting network $G_{rpn}$ and the discriminator $D_{rpn}$ for the domain alignment.

#### 3.3.2. Transferable region filtering

As shown in Figure 1, RPN extracts a number of region proposals with various anchors and then filters out *non-promising region proposals* according to the foreground probabilities, *i.e.* objectness scores. However, in the domain adaptation scenario, the predicted objectness scores on the target domain are unreliable since the objectness classifier in RPN-level is biased toward the source domain. As a result, some informative region proposals might have a low objectness score, which is removed during the proposal selection processes (*e.g.*, non-maximum suppression (NMS) and selecting top-$N$ proposals).

To alleviate this problem, we exploit the entropy of the output domain probability to estimate the transferability of each region proposal. Concretely, we calculate the entropy of each region proposal as $E_{i,r} = \mathcal{H}(p_{i,r}^{rpn})$, where high entropy indicates that the box is not distinguishable in terms of domains.

Next, we multiply the obtained domain transferability $E_{i,r}$ with the objectness score $o_{i,r} = G_{cls}(F_{rpn}(\tilde{f}_{i,L}))$, deriving the re-weighted objectness score $\tilde{o}_{i,r} = o_{i,r} \cdot E_{i,r}$. Afterwards, NMS is conducted to remove highly overlapping regions considering their re-weighted objectness scores. As illustrated in Figure 3, our filtering strategy preserves informative regions by involving the transferability in the selection process.

### 3.4. Dynamic Instance Sampling

For the instance-level domain adaptation, previous methods (Chen et al., 2020,1; Nguyen et al., 2020) select a fixed number of top-$N$ (*e.g.*, 300) region proposals of the target domain after non-maximum suppression (NMS). However, such the fixed-number strategy is undesirable in the adaptation scenario in that the region proposals of the target domain are unreliable especially in the early training steps, which eventually disturbs the adaptation process.

Hence, we propose *Dynamic Instance Sampling* (DIS), which dynamically adjusts the number of the selected region proposals from the target domain, *i.e.*, $N$, in the target region filtering step. Our intuition here is that only a small number of region proposals should be selected in the case when the *image-* and *RPN-level* alignment are insufficient. Besides, the number of region proposals should grow in a proportional way to the degree of the adaptation in the early components.

Firstly, we measure the Kullback Leibler (KL) divergence between foreground probabilities of regions from the RPN-level source and ones from the target domains. We utilize the divergence to estimate the relative perplexity of the model on the target domain compared to the source. The complement of the divergence is estimated as $1 - KL(o_n^s, o_n^t)$, where $KL(\cdot)$ denotes the KL divergence, $o_n^s$ and $o_n^t$ are the objectness score of selected top-$N$ region proposal in the source domain image and target domain one, respectively.

Moreover, we conjecture that the model should have a higher recall rate as the target feature gets closer to the source domain. From this intuition, we consider the domain probability of the image as an additional control parameter of $N$. The domain probability, i.e., the hardness score, is measured with an auxiliary domain discriminator $D_{dis}$ trained to distinguish the domain of multi-scale fused feature $M_i$. The final number of samples $N_{final}$ for the target image $x^t \in X_t$ is calculated as:

$$N_{final} = \lfloor N \times 0.5(D_{dis}(M_i^t) + (1 - KL(o_n^s, o_n^t))) \rfloor, \quad (8)$$

where $\lfloor \cdot \rfloor$ denotes the floor function. During training, $N_{final}$ gradually increases since the divergence $KL(o_n^s, o_n^t)$ decreases via gradient reversal layers in the overall network.

After Dynamic Instance Sampling (DIS), to align the instance-level distribution, we deploy an instance-level domain discriminator $D_{ins}$ with a GRL. In addition, inspired by (Saito et al., 2019), we concatenate image-level domain context vectors with instance-level feature vectors for regularizing the network. The instance-level adversarial loss is defined as follows:

$$\mathcal{L}_{adv}^{ins} = -\frac{1}{N_{final}} \sum_{i,n} \left( d_i \log p_{i,n}^{ins} + (1 - d_{i,n}) \log \left(1 - p_{i,n}^{ins}\right) \right), \quad (9)$$

where $p_{i,n}^{ins}$ is the probability of the instance-level domain discriminator $D_{ins}$ for the $n$-th region of interest sample in the $i$-th image.

### 3.5. Overall Training Objective

The overall loss function consists of the detection and adversarial losses, i.e., $\mathcal{L}_{adv} = \mathcal{L}_{adv}^{img} + \mathcal{L}_{adv}^{fus} + \mathcal{L}_{adv}^{rpn} + \mathcal{L}_{adv}^{ins}$. We train our framework with the following loss.

$$\mathcal{L}_{total} = \mathcal{L}_{det} + \mathcal{L}_{dis} + \lambda \cdot \mathcal{L}_{adv}, \quad (10)$$

where $\lambda$ is a trade-off parameter to balance the detector loss and our adversarial losses. And $\mathcal{L}_{dis}$ is the loss to train an auxiliary domain discriminator, $D_{dis}$. The network can be trained in an end-to-end manner. During inference, DIS and adversarial training are not conducted, whereas MUA and the transferable regions are estimated.

## 4. Experiments

In this section, we validate the effectiveness of our method in the domain adaptive object detection task on five benchmark datasets: Cityscapes (Cordts et al., 2016), FoggyCityscapes (Sakaridis, Dai and Van Gool, 2018), BDD100k (Yu et al., 2018), Sim10k (Johnson-Roberson et al., 2016) and KITTI (Geiger et al., 2013). We first describe the details of our experiments, and then analyze the quantitative and qualitative experimental results.

### 4.1. Dataset

**Cityscapes→FoggyCityscapes.** Cityscapes (Cordts et al., 2016) contains street scenes of 50 different cities in Germany. The training and test splits consist of 2,975 and 500 images with 8 different object categories, respectively. FoggyCityscapes (Sakaridis et al., 2018) is a synthetic dataset that is built by adding synthetic fogs to each image of Cityscapes. Following the convention of DAOD studies (Chen et al., 2018; Deng, Li, Chen and Duan, 2021; Xu, Zhao, Jin and Wei, 2020), we used the attenuation coefficient, i.e. foggy level, of 0.02 (Saito et al., 2019) for a fair comparison. In addition, we use the training set for model training and the testing set for evaluation following (Saito et al., 2019).

**Cityscapes→BDD100k.** BDD100k (Yu et al., 2018) is a large-scale driving dataset collected from various locations in the United States. It is composed of 70K training and 10K validation samples with 10 object categories. Due to the dynamic driving scenes of diverse conditions (e.g., weathers and locations), the dataset is known to be challenging. For evaluation, we utilize the Cityscapes training set for the source domain and the BDD100k training set of *daytime* for the target domain.

**Sim10k→Cityscapes.** Sim10k (Johnson-Roberson et al., 2016) is a virtual street scene dataset rendered from the Grand Theft Auto (GTA) game engine. It consists of 10,000 images with 58,701 box annotations of car objects. Following the protocol of (Chen et al., 2018; Saito et al., 2019), we used all images of Sim10K and Cityscapes training set during training. For the evaluation, we report the average precision of the common category, *car*, among both domains.

**KITTI→Cityscapes.** KITTI (Geiger et al., 2013) is collected from various streets and highways in Germany. KITTI contains 7,481 training images with bounding box annotations of cars. Since the image resolution difference between KITTI dataset and Cityscapes dataset, we resize the image from Cityscapes dataset by following (Zhu, Pang, Yang, Shi and Lin, 2019b).

### 4.2. Implementation Details

For a fair comparison, we adopt VGG-16 (Simonyan and Zisserman, 2014) based Faster R-CNN (Ren et al., 2015) with ROIAlign (He, Gkioxari, Dollár and Girshick, 2017) following previous works (Ren et al., 2015; Saito et al., 2019). We reported the performance trained after 70K iterations. The initial learning rate is set to $10^{-3}$ and adjusted to $10^{-4}$ after 50K iterations. For model optimization, we adopt the stochastic gradient descent (SGD) optimizer with the momentum of 0.9. We resize each image to have a vertical length of 600 pixels while preserving the aspect ratio. A single batch is composed of two images respectively for the source and target domains. We set $\lambda$ in Eq. 10 to 1 by default, except for Sim10K-to-Cityscapes, where $\lambda$ is set to 0.1. Our code is implemented with PyTorch (Paszke, Gross, Chintala, Chanan, Yang, DeVito, Lin, Desmaison, Antiga

**Table 2**
Quantitative results on Cityscapes→FoggyCityscapes. We report the average precision (%) for all categories on the target domain.

| Method | Person | Rider | Car | Truck | Bus | Train | Motorbike | Bicycle | mAP |
|---|---|---|---|---|---|---|---|---|---|
| Source only (Ren et al., 2015) | 24.1 | 33.1 | 34.3 | 4.1 | 22.3 | 3.0 | 15.3 | 26.5 | 20.3 |
| DA-Faster (Chen et al., 2018) | 25.0 | 31.0 | 40.5 | 22.1 | 35.3 | 20.2 | 20.0 | 27.1 | 27.6 |
| SCDA (Zhu et al., 2019a) | 33.5 | 38.0 | 48.5 | 26.5 | 39.0 | 23.3 | 28.0 | 33.6 | 33.8 |
| SW-Faster (Saito et al., 2019) | 29.9 | 42.3 | 43.5 | 24.5 | 36.2 | 32.6 | 30.0 | 35.3 | 34.3 |
| ATF (He and Zhang, 2020) | 34.6 | 47.0 | 50.0 | 23.7 | 43.3 | 38.7 | 33.4 | 38.8 | 38.7 |
| HTCN (Chen et al., 2020) | 33.2 | 47.5 | 47.9 | 31.6 | 47.4 | 40.9 | 32.3 | 37.1 | 39.7 |
| RDA (Huang et al., 2021) | 37.4 | 46.4 | 48.3 | 32.6 | 46.5 | 39.3 | 32.3 | 35.3 | 39.8 |
| MEAA (Nguyen et al., 2020) | 34.2 | 48.9 | 52.4 | 30.3 | 42.7 | 46.0 | 33.2 | 36.2 | 40.5 |
| RPA (Zhang, Wang and Mao, 2021) | 33.6 | 43.8 | 49.6 | 32.9 | 45.5 | 46.0 | 35.7 | 36.8 | 40.5 |
| DSS (Wang, Zhang, Zhang, Li, Xia, Zhang and Liu, 2021) | **42.9** | **51.2** | 53.6 | 33.6 | 49.2 | 18.9 | 36.2 | 41.8 | 40.9 |
| UaDAN (Guan et al., 2021) | 36.5 | 46.1 | 53.6 | 28.9 | 49.4 | 42.7 | 32.3 | 38.9 | 41.1 |
| UMT (Deng et al., 2021) | 33.0 | 46.7 | 48.6 | **34.1** | 56.5 | 46.8 | 30.4 | 37.3 | 41.7 |
| MeGA-CDA (VS et al., 2021) | 37.7 | 49.0 | 52.4 | 25.4 | 49.2 | **46.9** | **34.5** | **39.0** | 41.8 |
| Ours | 34.0 | 47.8 | 52.5 | 32.5 | 51.7 | 46.5 | **34.5** | 38.4 | **42.2** |
| Oracle | 37.2 | 48.2 | 52.7 | 35.2 | 52.2 | 48.5 | 35.3 | 38.8 | 43.5 |

**Table 3**
Quantitative results on Cityscapes→BDD100K. We report the average precision (%) for all categories on the target domain. * denotes the reproduced results.

| Method | Person | Rider | Car | Truck | Bus | Motorbike | Bicycle | mAP |
|---|---|---|---|---|---|---|---|---|
| Source only (Ren et al., 2015) | 26.9 | 22.1 | 44.7 | 17.4 | 16.7 | 17.1 | 18.8 | 23.4 |
| DA-Faster (Chen et al., 2018) | 29.4 | 26.5 | 44.6 | 14.3 | 16.8 | 15.8 | 20.6 | 24.0 |
| SW-Faster (Saito et al., 2019) | 30.2 | 29.5 | 45.7 | 15.2 | 18.4 | 17.1 | 21.2 | 25.3 |
| MEAA* (Nguyen et al., 2020) | 29.2 | 30.6 | 43.5 | 20.2 | 21.5 | 18.8 | 23.5 | 26.5 |
| ICR-CCR (Xu et al., 2020) | 31.4 | 31.3 | **46.3** | 19.5 | 18.9 | 17.3 | 23.8 | 26.9 |
| Ours | **34.0** | **33.4** | 45.6 | **21.2** | **25.6** | **19.9** | **29.3** | **29.9** |

and Lerer, 2017). We conduct all the experiments on a single RTX2080Ti.

### 4.3. Comparison with state-of-the-art methods
#### 4.3.1. Quantitative comparison

We validate the effectiveness of the proposed methods on various unsupervised domain adaptive object detection benchmarks. As shown in Table 2, our methods outperform previous methods, achieving a new state-of-the-art performance of 42.2% on Cityscapes→FoggyCityscapes.

In addition, our model also surpasses the previous methods on the Cityscapes→BDD100K benchmark by a large margin of 3.0%, as shown in Table 3. Notably, the detection performance on various size object classes including large object classes (*e.g.*, Truck and Bus) is largely improved despite the challenging nature of BDD100K.

We also compare our method with previous approaches on three car-only adaptation scenarios, *i.e.*, Sim10k → Cityscapes and KITTI ↔ Cityscapes. As shown in Table 4, our method outperforms the previous state-of-the-arts by achieving the AP of 44.2% and 47.1% for Sim10k → Cityscapes, KITTI→Cityscapes scenarios. Moreover, our method provides detection performance of 75.2% for Cityscapes → KITTI scenarios, which is comparable achievement to the current state-of-the-art MeGA-CDA (VS et al., 2021) which scored 75.5%.

#### 4.3.2. Qualitative comparison

We provide qualitative results that are adapted from Cityscapes to FoggyCityscapes and BDD100K in Figure 4. For comparison, we also present the results from one of the state-of-the-arts, MEAA (Nguyen et al., 2020), which leverages semantic guidance by measuring the domain uncertainty of local feature representation in image-level. In the visualizations, heavy fogs in the target domain disturb accurate recognition, especially distant objects as shown in Figure 4a. MEAA better detects objects than the baseline with the help of the local feature domain uncertainty map but fails to capture some objects due to the noisy activation (see the $2^{nd}$ and $4^{th}$ rows). For instance, a large bus in the first sample is partially detected with low confidence, and distant cars are not discovered in the second image. On the contrary, our method captures almost all the objects well. To be specific, the large bus is accurately detected since the Transferable Region Proposal Network (TRPN) effectively minimizes the domain gap of the objectness distributions from anchors with various shapes. In addition, the Multi-scale-aware Uncertainty Attention (MUA) provides local and global image information, enabling the model to precisely capture small cars in the second sample without false positives.

### 4.4. Analysis
#### 4.4.1. The effectiveness of each component

Following the previous studies (Chen et al., 2020; Nguyen et al., 2020), we utilize both SW-Faster (Saito et al., 2019)

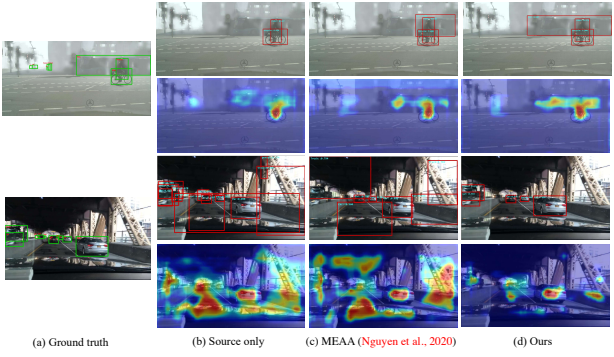

**Figure 4:** Qualitative results on Cityscapes→FoggyCityscapes (top) and Cityscapes→BDD100K (bottom).

**Table 4**
Quantitative results with SIM10K→City (1st row), KITTI→City (2nd row) and City→KITTI (3rd row). We report the average precision (%) on the "car" class.

| Model | SIM10K→City | KITTI→City | City→KITTI |
|---|---|---|---|
| Source-only (Ren et al., 2015) | 34.6 | 30.2 | 53.5 |
| DA-Faster (Chen et al., 2018) | 38.9 | 38.5 | 64.1 |
| SW-DA (Saito et al., 2019) | 40.1 | 37.9 | 71.0 |
| MEAA (Nguyen et al., 2020) | 42.0 | - | - |
| ATF (He and Zhang, 2020) | 42.8 | 42.1 | 73.5 |
| SC-DA (Zhu et al., 2019b) | 43.1 | 42.5 | - |
| UMT (Deng et al., 2021) | 43.1 | - | - |
| DSS (Wang et al., 2021) | 44.5 | 42.7 | - |
| CT-DA (Zhao et al., 2020) | 44.5 | 43.6 | - |
| RPA (Zhang et al., 2021) | 45.7 | - | - |
| IFAN (Zhuang, Han, Huang and Scott, 2020) | 46.9 | - | - |
| MeGA-CDA (VS et al., 2021) | 44.8 | 43.5 | 75.5 |
| Ours | 47.1 | 44.2 | 75.2 |

with additional domain discriminators for middle features in image-level and domain discriminator instance-level inspired by DA-Faster (Chen et al., 2018) as our baseline domain adaptation method. Based on this method, we conduct ablation studies to validate the effectiveness of individual proposed components: Multi-scale-aware Uncertainty Attention (MUA), Transferable Region Proposal Network (TRPN), and Dynamic Instance Sampling (DIS). As shown in Table 5, each component contributes to the performance improvement. Specifically, we observe that the transferable region guidance from MUA benefits the performance for all the object categories, improving the mAP score by 0.8% (see the 3rd row). Meanwhile, our TRPN brings a large performance improvement of 1.8% (see the 6th row), indicating that the region proposals should be selected regarding their transferability as well as the objectness scores. Furthermore, DIS enhances the adaptation performance effectively by incorporating the extent of the domain gap into the region sampling process with an improvement of 1.0% (refer to the 4th row). To summarize, all the proposed components are beneficial for adaptation and complementary to each other, which is consistent with our motivation that the individual components in the object detection pipeline are interdependent and closely related to each other.

### 4.4.2. Analysis on Dynamic Instance Sampling

We experimentally validate the motivation of the Dynamic Instance Sampling (DIS) that the recall rate should increase as the target feature gets close to the source domain. In Figure 5, we display a scatter plot of 500 images from the target domain, where X-axis and Y-axis are the domain probability and the recall rate, respectively.

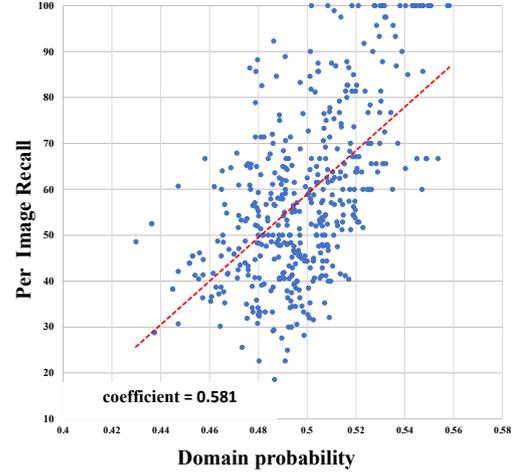

**Figure 5:** Analysis on the correlation between the domain probability and the recall rate.

To investigate the relationship between the recall rate and the domain probability, we measure the Pearson correlation coefficient. As a result, it is noticed that the correlation between the domain probability and the recall rate is strong with its coefficient of 0.58, which is consistent with our motivation.

Our DIS calibrates the number of the target region proposals to be selected by considering the output domain probability of the target domain image and divergence between foreground probabilities of regions from source and target domain. In Figure 6, we illustrate the tendency of $N_{final}$ of Dynamic Instance Sampling (DIS) during training on Cityscapes → FoggyCityscapes, where $N$ in Eq.8 is 300. $N_{final}$ initially starts from 150, then drastically gets close to 250 after 5K iterations. Afterward, $N_{final}$ continuously oscillates reflecting the adaptation progress, and finally converges to 280.

Additionally, we qualitatively analyze the effect of DIS in Figure 7. As shown in Figure 7b, the conventional top-$N$ sampling strategy selects unpropitious region proposals including backgrounds due to the large domain gap. On the contrary, DIS automatically adjusts the number of proposals, so background regions are excluded while foreground regions are selected.

### 4.5. Additional Visualization Results
#### 4.5.1. Multi-scale-aware Uncertainty Attention

Figure 8 qualitatively compares the uncertainty attention estimated from different feature scales in image-level: low-features (*Conv3*), global-features (*Conv5*), foreground prediction confidence from RPN (Guan et al., 2021), and the proposed multi-scale fusion features. For visualization, we draw attention maps on the original inputs from the Foggy-Cityscapes test set using the jet colormap, where highly activated regions are close to red, otherwise to blue. Compared to the single-scale features, MUA better captures the transferable regions, including all of the foreground objects from small ones (e.g., distant cars) to large ones (e.g., buses and trucks) despite their blurred appearance. Consequently, MUA effectively provides the transferability guidance to the subsequent modules, i.e., RPN and the detection head, and also enhances the discriminability by preventing negative transfer caused by background regions.

**Table 5**
Ablation studies on Cityscapes→FoggyCityscapes. The backbone network is VGG-16 (Simonyan and Zisserman, 2014). Following previous works (Chen et al., 2020; Nguyen et al., 2020), we set our baseline as SW-Faster (Saito et al., 2019) with additional domain alignment loss (Chen et al., 2018) for image-level and instance-level features.

| Method | MUA | TRPN | DIS | Person | Rider | Car | Truck | Bus | Train | Motorbike | Bicycle | mAP | Gain |
|---|---|---|---|---|---|---|---|---|---|---|---|---|---|
| Source only | - | - | - | 24.1 | 33.1 | 34.3 | 4.1 | 22.3 | 3.0 | 15.3 | 26.5 | 20.3 | - |
| Baseline | - | - | - | 32.5 | 43.6 | 44.9 | 28.4 | 41.8 | 40.0 | 30.2 | 36.3 | 37.3 | - |
| Ours | ✓ | - | - | 32.8 | 44.7 | 45.3 | 28.8 | 44.2 | 41.0 | 31.5 | 36.8 | 38.1 | 0.8 |
| | - | - | ✓ | 32.1 | 45.8 | 46.9 | 29.1 | 43.4 | 40.7 | 32.7 | 36.1 | 38.3 | 1.0 |
| | ✓ | - | ✓ | 33.1 | 46.2 | 47.0 | 29.6 | 47.2 | 42.4 | 32.9 | 37.0 | 39.4 | 2.1 |
| | - | ✓ | - | 32.7 | 46.1 | 46.7 | 30.5 | 46.8 | 42.1 | 31.3 | 36.9 | 39.1 | 1.8 |
| | ✓ | ✓ | - | 32.9 | 46.3 | 48.9 | 31.0 | 49.1 | 45.8 | 33.0 | 37.2 | 40.4 | 3.1 |
| | - | ✓ | ✓ | 33.4 | 47.0 | 50.1 | 31.8 | 50.6 | 44.7 | 33.2 | 38.0 | 41.1 | 3.8 |
| | ✓ | ✓ | ✓ | **34.0** | **47.8** | **52.5** | **32.5** | **51.7** | **46.5** | **34.5** | **38.4** | **42.2** | **4.9** |
| Oracle | - | - | - | 37.2 | 48.2 | 52.7 | 35.2 | 52.2 | 48.5 | 35.3 | 38.8 | 43.5 | - |

**Table 6**
Ablation studies on Cityscapes→FoggyCityscapes. The backbone network is ResNet-101 (He, Zhang, Ren and Sun, 2016).

| Method | MUA | TRPN | DIS | Person | Rider | Car | Truck | Bus | Train | Motorbike | Bicycle | mAP | Gain |
|---|---|---|---|---|---|---|---|---|---|---|---|---|---|
| Source only | - | - | - | 26 | 37.2 | 35.7 | 18 | 26.1 | 6.1 | 22.6 | 28.3 | 25.0 | - |
| Baseline | - | - | - | 27.8 | 36.9 | 36.2 | 22.2 | 34 | 10.2 | 22.8 | 29.4 | 27.4 | - |
| Ours | ✓ | - | - | 29.5 | 42.1 | 40.8 | 24.4 | 37.7 | 20.5 | 25.5 | 33.2 | 31.7 | 4.3 |
| | - | - | ✓ | 27.1 | 37.5 | 35.9 | 23.5 | 35.9 | 14.8 | 23.5 | 29.7 | 28.5 | 1.1 |
| | ✓ | - | ✓ | 29.8 | 43.1 | 40.2 | 24.9 | 38.4 | 28.2 | 25.3 | 33.7 | 33.0 | 6.5 |
| | - | ✓ | - | 28.3 | 41.7 | 41.3 | 23.3 | 39.1 | 15.6 | 21.4 | 33.9 | 30.6 | 3.2 |
| | ✓ | ✓ | - | 30.1 | 42.8 | 42.2 | 26.0 | 38.9 | 30.3 | 25.9 | 34.6 | 33.9 | 6.5 |
| | - | ✓ | ✓ | 30.4 | 43.6 | 42.9 | 25.3 | 39.5 | 22.9 | 24.5 | 34.2 | 32.9 | 5.5 |
| | ✓ | ✓ | ✓ | **31.6** | **44.2** | **43.1** | **27.6** | **40.9** | **32.2** | **26.5** | **34.9** | **35.1** | **7.7** |
| Oracle | - | - | - | 31.5 | 44 | 45.1 | 33.7 | 54.1 | 44.5 | 33.3 | 29.7 | 39.5 | - |

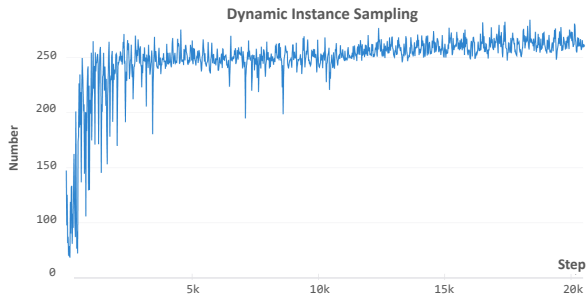

**Figure 6:** The tendency of $N_{final}$ of Dynamic Instance Sampling (DIS) during training on Cityscapes→FoggyCityscapes. X-axis and Y-axis denote the final number of samples $N_{final}$ and the elapsed iterations, respectively.

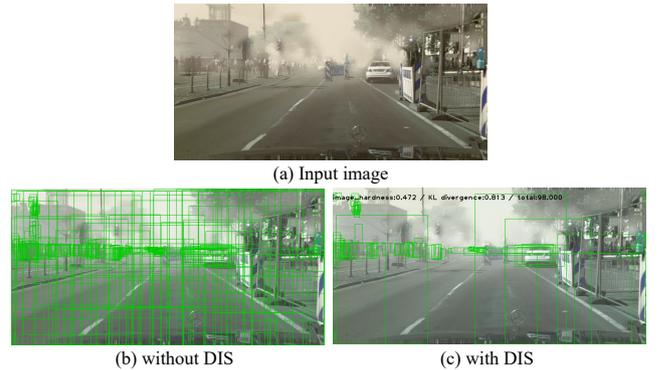

**Figure 7:** Top-$N$ region proposal visualization. (a) an input image (b) the results without DIS (i.e., $N_{final}$ is fixed to 300) (c) DIS adjusts $N_{final}$ to be 98.

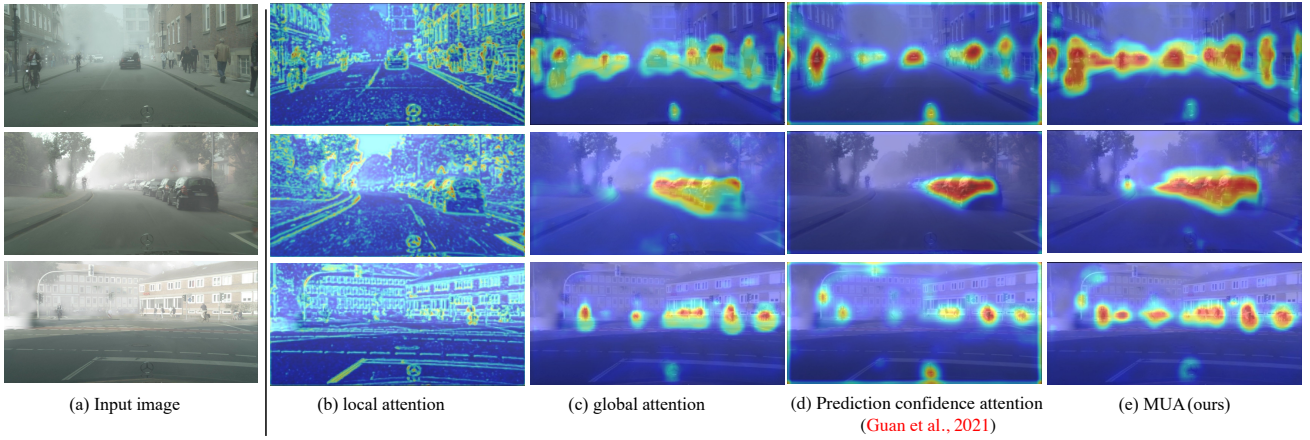

**Figure 8:** Visualization of attention maps on Cityscapes→FoggyCityscapes. From left to right, the columns depict (a) input target images, (b) attention maps from local features, (c) those from global features, (d) foreground prediction confidence attention (Guan et al., 2021), and (e) those from the proposed MUA, respectively.

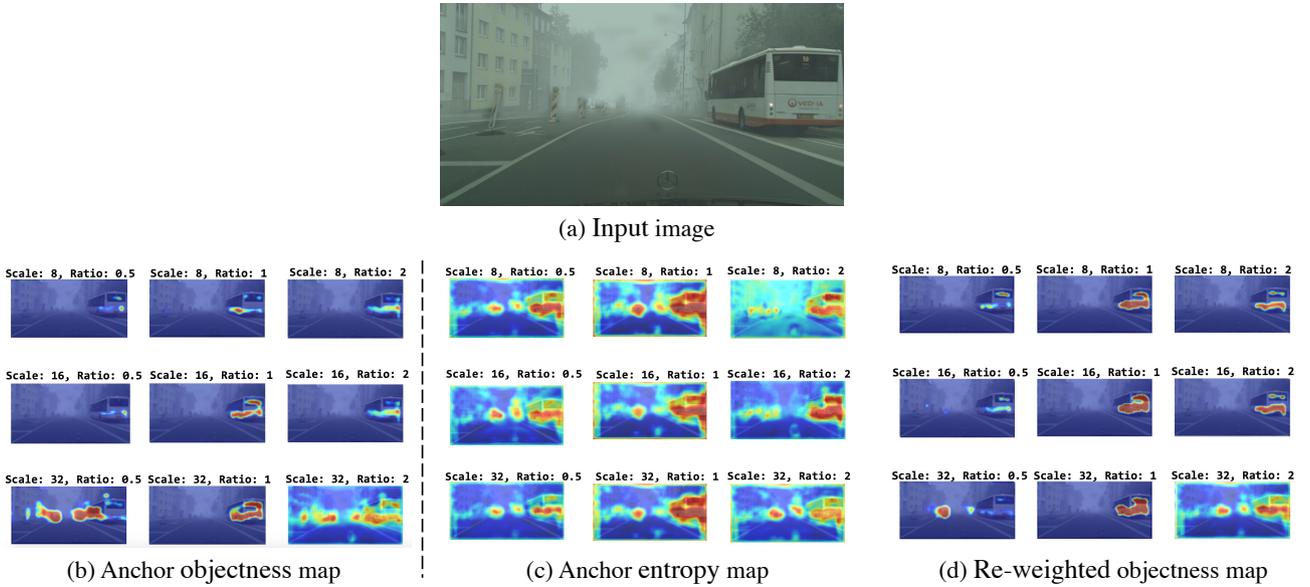

**Figure 9:** Visualization of the effectiveness of TRPN. (a) an input image, (b) objectness score maps for different anchors of the vanilla RPN, (c) the entropy maps for the anchors, (d) the re-weighted objectness score maps for the anchors of TRPN.

### 4.5.2. Transferable Region Proposal Network

In Figure 9, we analyze the Transferable Region Proposal Network (TRPN) with the qualitative results. Without RPN-level adaptation, objectness score maps obtained from anchors are insufficient to distinguish foregrounds from the backgrounds due to the large domain gap (Figure 9b). On the other hand, the entropy map of the domain probability can provide the guidance of transferable regions, as shown in Figure 9c. Especially, we can observe that foreground regions are usually regarded to be transferable, whereas backgrounds are excluded. Consequently, TRPN exploits the re-weighted objectness score maps, and therefore better discriminates the foreground regions by simultaneously taking into account both the foreground score and the transferability.

## 4.6. Additional Qualitative Detection Results

In this section, we provide additional qualitative results on different domain adaptation scenarios. Figure 10 shows results of SIM10k-to-Cityscapes (top) and KITTI-to-Cityscapes (bottom). Besides, we illustrate qualitative results on Cityscapes-to-FoggyCityscapes (top) and Cityscapes-to-BDD100K (bottom) in Figure 11. In every domain adaptation scenario, our method effectively reduces false positives, while greatly empowering the detection ability to accurately capture objects regardless of their size (refer to the 4$^{th}$ and 8$^{th}$ rows).

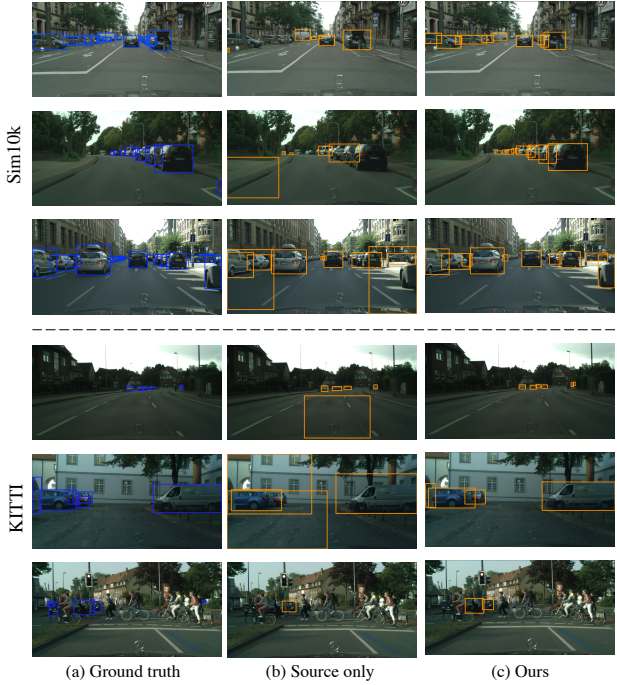

Figure 10: Detection results on the car detection task: (top) SIM10k→Cityscapes and (bottom) KITTI→Cityscapes. From left to right, the columns show (a) the ground truth, (b) the results of the source only model, and (c) the proposed method, respectively.

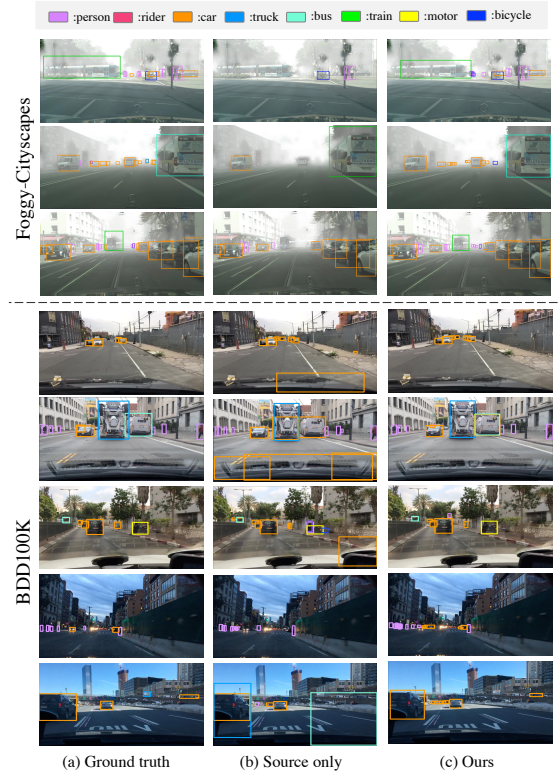

Figure 11: Detection results on two adaptation scenarios: (top) Cityscapes→FoggyCityscapes (Sakaridis et al., 2018) and (bottom) Cityscapes→BDD100k (Yu et al., 2018). From left to right, the columns show (a) the ground truth, (b) the results of source only model, and (c) the proposed method, respectively.

### 4.7. Computation time

In this section, we provide computation time comparison with one of the state-of-the-arts, MEAA (Nguyen et al., 2020). We measured the time during training and inference in the city-to-foggy scenario with a single GTX2080Ti. More specifically, we have measured the training time for 1 epoch and the inference time for the entire test set. As a result, our model is shown to be more efficient (train: 53.29 min | test :43.85 sec) compared to MEAA (Nguyen et al., 2020) (train: 55.9 min | test: 47.05 sec ).

## 5. Conclusion

In this work, we proposed a novel framework including three components for individual level adaptation: Multi-scale-aware Uncertainty Attention (MUA), Transferable Region Proposal Networks (TRPN), and Dynamic Instance Sampling (DIS). By carefully handling the inter-relation of the components of object detector, our methods effectively reduced the domain gap and maximize transferability as well as discriminability for domain adaptive object detection. Through extensive experiments, we validated the effectiveness of our method by achieving a new state-of-the-art performance in various domain adaptation scenarios.

## Acknowledgments

This work was supported by Institute of Information & communications Technology Planning & Evaluation (IITP) grant funded by the Korea government(MSIT) (No. 2018-0-00769: Neuromorphic Computing Software Platform for Artificial Intelligence Systems and No. 2020-0-01361: Artificial Intelligence Graduate School Program (YONSEI UNIVERSITY)).

## CRediT authorship contribution statement

**Mirae Do:** Conceptualization, Methodology, Software, Visualization, Investigation, Writing- Original draft. **Seogkyu Jeon:** Methodology, Investigation, Writing- Original draft preparation, Writing- Reviewing and Editing. **Pilhyeon Lee:** Visualization, Investigation, Writing- Original draft preparation, Writing- Reviewing and Editing. **Kibeom hong:** Conceptualization, Writing- Original draft preparation, Writing- Reviewing and Editing. **Yu-seung Ma:** Conceptualization, Supervision. **Hyeran Byun:** Supervision, Method Improvement, Writing- Reviewing and Editing.

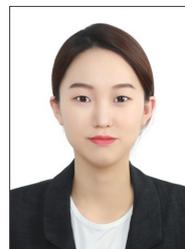

**Mirae Do** is currently a M.S student in Artificial Intelligence at Yonsei University, Seoul, Korea. Her received the B.S. degree in Computer Science from Duksung Women's University, Seoul, Korea. Her research interests include domain adaptation and object detection.

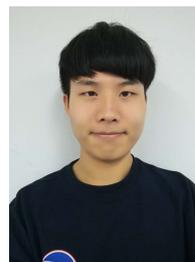

**Seogkyu Jeon** is currently a Ph.D. student in Computer Science at Yonsei University, Seoul, Korea. He received the B.S. degree in the Department of Software from Konkuk University, Seoul, Korea. His research interests include domain adaptation, domain generalization, and image-to-image translation.

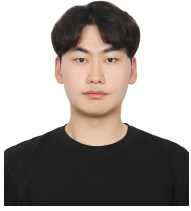

**Pilhyeon Lee** is currently a Ph.D. student in Computer Science at Yonsei University, Seoul, Korea. He received the B.S. degree in Computer Science and Engineering from Chung-Ang University, Seoul, Korea. His research interests include video analysis, weakly-supervised learning, and representation learning.

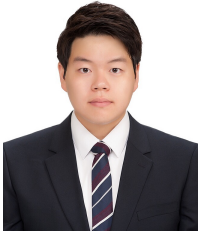

**Kibeom Hong** is currently a Ph.D. student in Computer Science at Yonsei University, Seoul, Korea. He received the B.S. degree in Computer Science from Yonsei University, Seoul, Korea. His research interests include generative adversarial networks, generative models, and neural style transfer.

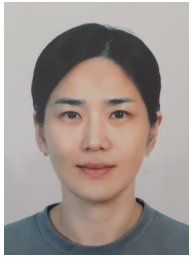

**Yu-seung Ma** received the B.S., M.S., and Ph.D. degrees in computer science from Korea Advanced Institute of Science and Technology (KAIST), Korea, in 1998, 2000 and 2005, respectively. In 2005, she joined in the Electronics and Telecommunications Research Institute (ETRI), Korea. Her research interests include program testing, machine learning, and neuromorphic computing.

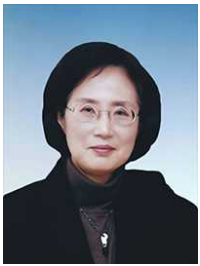

**Hyeran Byun** is currently a professor of Computer Science at Yonsei University. She was an Assistant Professor at Hallym University, Chooncheon, Korea, from 1994 to 1995. She served as a non executive director of National IT Indus-try Promotion Agency (NIPA) from Mar. 2014 to Feb. 2018. She is a member of National Academy Engineering of Korea. Her research interests include computer vision, image and video processing, deep learning, artificial intelligence, machine learning, and pattern recognition. She received the B.S. and M.S. degrees in mathematics from Yonsei University, Seoul, Korea, and the Ph.D. degree in computer science from Purdue University, West Lafayette, IN, USA.